%% file: main.tex
\documentclass[letterpaper, conference]{IEEEtran}

\IEEEoverridecommandlockouts                              % This command is only needed if 
\usepackage{subcaption}
\usepackage{hyperref}       % hyperlinks
\usepackage{booktabs}       % professional-quality tables
\usepackage{placeins}
\usepackage[nocompress]{cite}
\usepackage{graphicx}
\usepackage[acronym,shortcuts]{glossaries}
\usepackage{comment}
\usepackage{enumerate}
\usepackage[capitalise]{cleveref}
\usepackage[table,xcdraw,dvipsnames]{xcolor}
\usepackage{siunitx}
\usepackage{bm} % Package for bold symbols in math mode
\usepackage{amssymb} % Include the amssymb package for \mathbb and other math symbols
\usepackage{amsmath}
\usepackage[english]{babel}
\usepackage{multirow}
\usepackage{pifont}
\usepackage{algorithm}
\usepackage{algpseudocode}
\usepackage{cuted}
\usepackage[bb=dsserif]{mathalpha}

\definecolor{lightgray}{gray}{0.9}

\def\BibTeX{{\rm B\kern-.05em{\sc i\kern-.025em b}\kern-.08em
    T\kern-.1667em\lower.7ex\hbox{E}\kern-.125emX}}
\title{\LARGE \bf \vspace{6mm}
{\namealgo}: A Residual Method for Zero-Shot Real-World \\Autonomous Racing on Scaled Platforms 
}

\author{Edoardo Ghignone\IEEEauthorrefmark{1}, Nicolas Baumann\IEEEauthorrefmark{1}, Cheng Hu\IEEEauthorrefmark{2}, Jonathan Wang\IEEEauthorrefmark{1}\\ Lei Xie\IEEEauthorrefmark{2}, Andrea Carron \IEEEauthorrefmark{3}, and Michele Magno\IEEEauthorrefmark{1} \\% <-this % stops a space
\thanks{\IEEEauthorrefmark{1}Edoardo Ghignone, Nicolas Baumann, Jonathan Wang, and Michele Magno are associated with the Center for Project-Based Learning, D-ITET, ETH Zurich.}%
\thanks{\IEEEauthorrefmark{2}Cheng Hu and Lei Xie are associated with the Department of Control Science and Engineering, Zhejiang University.}
\thanks{\IEEEauthorrefmark{3}Andrea Carron is associated with the Institute for Dynamic Systems and Control, D-MAVT, ETH Zürich.}
\thanks{\emph{(Edoardo Ghignone, Nicolas Baumann, and Cheng Hu contributed equally to this work.) (Corresponding author: Edoardo Ghignone.)}}
}

\begin{document}

\include{acronyms}
\include{keywords}

\maketitle

\begin{strip}
\vspace{-3.25cm}
\centering
\includegraphics[width=\textwidth, trim={24cm 0cm 20cm 7cm}, clip]{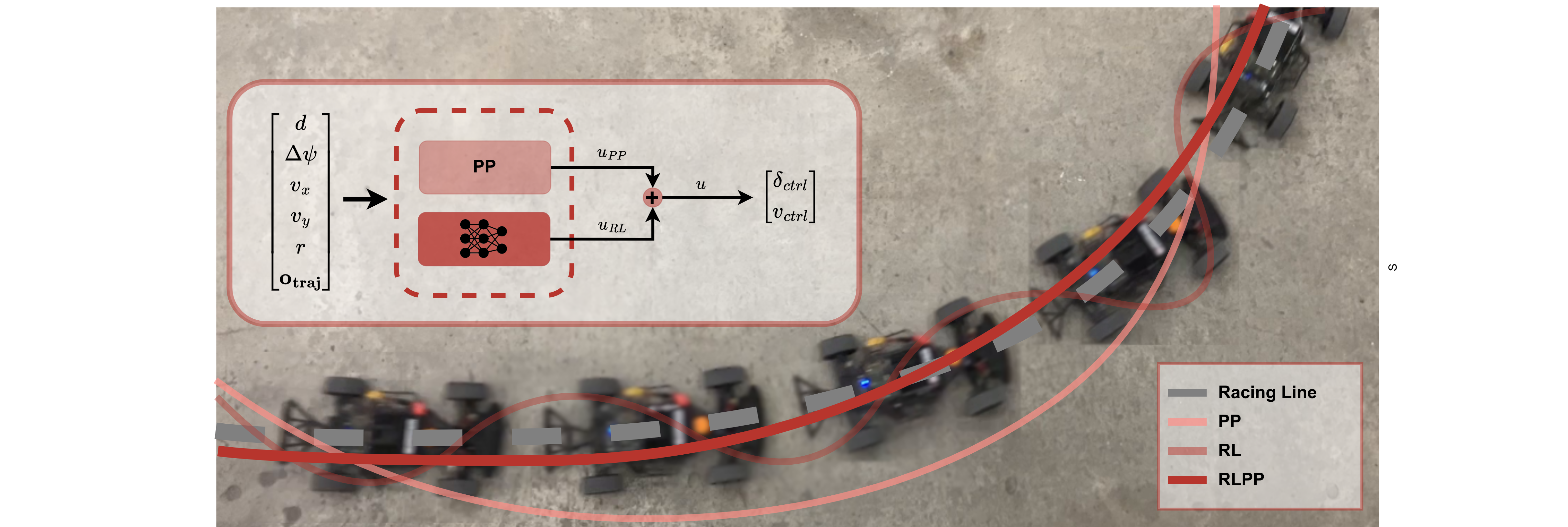}
\captionof{figure}{\textbf{\namealgo}: the residual structure, summarized in the picture, allows for integrating an \acrshort{rl} network with a classical controller, namely \acrshort{pp}. Such architecture retains the tuning capabilities of the traditional method with the performance increase of the data-driven \acrshort{rl} approach, hence enabling zero-shot deployment on a real-world platform and lap time improvement, without requiring system identification techniques as for \acrshort{sota} approaches.}
\label{fig:graphical_abstract}
\vspace{-0.275cm}
\end{strip}

\thispagestyle{empty}
\pagestyle{empty}

%%%%%%%%%%%%%%%%%%%%%%%%%%%%%%%%%%%%%%%%%%%%%%%%%%%%%%%%%%%%%%%%%%%%%%%%%%%%%%%%
\begin{abstract}
Autonomous racing presents a complex environment requiring robust controllers capable of making rapid decisions under dynamic conditions. While traditional controllers based on tire models are reliable, they often demand extensive tuning or system identification. \gls{rl} methods offer significant potential due to their ability to learn directly from interaction, yet they typically suffer from the \sim2real gap, where policies trained in simulation fail to perform effectively in the real world. In this paper, we propose {\namealgo}, a residual \gls{rl} framework that enhances a \gls{pp} controller with an \gls{rl}-based residual. This hybrid approach leverages the reliability and interpretability of \gls{pp} while using \gls{rl} to fine-tune the controller's performance in real-world scenarios. Extensive testing on the F1TENTH platform demonstrates that {\namealgo} improves lap times of the baseline controllers by up to \qty{6.37}{\percent}, closing the gap to the \gls{sota} methods by more than \qty{52}{\percent} and providing reliable performance in zero-shot real-world deployment, overcoming key challenges associated with the \sim2real transfer and reducing the performance gap from simulation to reality by more than 8-fold when compared to the baseline \gls{rl} controller. The {\namealgo} framework is made available as an open-source tool, encouraging further exploration and advancement in autonomous racing research.
The code is available at: \url{www.github.com/forzaeth/rlpp}.
\end{abstract}

% \begin{IEEEkeywords}
%     Reinforcement Learning, Field Robots, Wheeled Robots
% \end{IEEEkeywords}

%%%%%%%%%%%%%%%%%%%%%%%%%%%%%%%%%%%%%%%%%%%%%%%%%%%%%%%%%%%%%%%%%%%%%%%%%%%%%%%%
\setlength{\tabcolsep}{3pt}

\section{Introduction} \label{sec:intro}
\gls{rl} offers a promising solution for racing setups, being able to achieve super-human performance in driving simulators \cite{fuchs2021superhuman}, and similarly being able to reach the level of human professionals in the setting of drone racing \cite{champion_level}. 
For this reason, different works have tried to replicate these feats in the domain of scaled autonomous car racing, such as in \cite{chisari2021learning}, where an \gls{rl} agent reaches levels of a \gls{sota} \gls{mpc} baseline after finetuning on the real platform.
However, such results are obtained with an external motion capture system: this significantly simplifies the task at hand when compared to fully-onboard robotic platforms such as F1TENTH \cite{okelly2020f1tenthsim}, where the more realistic setting constrains the car to noisy localization and state estimation.
In this case, the performance of \gls{rl} techniques is lagging behind when compared to \gls{sota} traditional control techniques, such as in \cite{map}. 
 \cite{brunnbauer2022latent} presents an interesting \gls{rl} example, where the Dreamer architecture \cite{hafner2020dreamer} is used to show that model-based \gls{rl} algorithms generalize well to multiple tracks when compared to the model-free counterparts in simulation. However, when considering deployment on the physical platform, model mismatches are witnessed by the authors, specifically with the appearance of a bang-bang steering behavior, that limits the maximum speed of the car when compared to simulation.
Similarly, in \cite{Ghignone2023tcdriver}, a hybrid method unifying traditional planners and an \gls{rl} controller is shown, but deployment on the real platform is constrained to final velocities slower than those during training due to the encountered steering misbehavior. Finally, the behavior of different \gls{rl} algorithms on the F1TENTH platform is compared in \cite{EVANS2023comparing}, where end-to-end algorithms are shown to be transferable from simulation to reality at speeds up to \SI{5}{\metre \per \second}, but without demonstrating repeated laps, and highlighting the need for further study in the \sim2real transfer domain.

This work aims to advance Sim2Real techniques by exploring a different architecture of reinforcement learning controllers, namely those of residual \gls{rl} policies. Introduced with \cite{joahnnink2019residual} for a robot manipulator, this method consists of complementing a traditional type of controller with a \gls{rl} policy, relinquishing part of the control authority and simplifying the task for the \gls{ml}-based agent. This technique was further deployed on the F1TENTH simulator  \cite{trumpp2023residual, trumpp2024racemop}, attaining lap time improvement and overtaking behavior on different tracks.
However, we argue such architecture can be transferred to the physical platform and can help to bridge the \sim2real gap, by retaining the tuning simplicity of the baseline algorithm while allowing for a data-driven enhancement.

This paper presents {\namealgo}, a residual \gls{rl} controller that enhances the baseline performance of a \gls{pp} controller with a residual action, using only simulator experience. Differently from work that applied the residual structure with controllers using tire knowledge \cite{bang2024rlaugmentedmpcframeworkagile}, we apply the technique to a tire-free traditional method, simplifying the overall procedure by removing the need for system identification.
Furthermore, we deploy our system on the F1TENTH platform, where the full perception-planning-control is run on board and no external communication or computing budget is provided (see \Cref{tab:rw} for a comparison).

The main contributions of this paper are as follows:
\begin{enumerate}[I]
    % \item We present {\namealgo}  that shows behavior transfer of the learned agent on a real platform, highlighting what calibration parameters are retained by the hybrid architecture, and arguing that such parameters are needed in cases where the simulator is particularly simple, such as the F1TENTH one; To the best of our knowledge, we are the first to demonstrate how an \gls{rl} algorithm reaches a performance level close to the top performing \gls{ar} methods on a real embedded platform such as the F1TENTH platform.
    \item 
    We present {\namealgo}, a residual \gls{rl} method combining a traditional controller with a \gls{ml}-driven policy. The presented approach retains few calibration parameters, simplifying the behavior transfer across the \sim2real gap. 
    To the best of our knowledge, this is the first demonstration of an \gls{rl} algorithm achieving near top-level performance compared to leading \gls{ar} methods on an embedded platform like F1TENTH.
    \item We extensively compare the proposed method with tire-free and tire-aware \gls{sota} \gls{ar} controllers, showcasing specifically how {\namealgo} can improve the performance of a baseline controller.
    \item We provide the evaluated implementation (\gls{rl} environment, training procedures, and final controller) as an open-source modular addition to the software architecture presented in \cite{forzaeth}.
\end{enumerate}

\begin{table}[bt]
    \centering
    \vspace{0.15cm}
    \label{tab:rw}
    \begin{tabular}{lccc}
        \toprule
         Related Work & Fully Onboard & Multiple Laps & Max Speed \\
         \midrule
        Chisari et al. \cite{chisari2021learning}  & No & \textbf{Yes} & \textbf{\qty[text-series-to-math]{3}{\metre \per \second}} $\dagger$\\
        Brunnbauer et al. \cite{brunnbauer2022latent} & \textbf{Yes} & \textbf{Yes} & N/A \\
        TC-Driver \cite{Ghignone2023tcdriver} & \textbf{Yes} & \textbf{Yes} & \SI{3}{\metre \per \second} \\
        TAL \cite{EVANS2023comparing} & \textbf{Yes} & No & \SI{5}{\metre \per \second}\\
        \textbf{Ours} & \textbf{Yes} & \textbf{Yes} & \textbf{\qty[text-series-to-math]{5.4}{\metre \per \second}} \\
        \bottomrule
    \end{tabular}
        \caption{Comparison of related works on \gls{rl} controllers for scaled \gls{ar}. The used platform is F1TENTH, except for \cite{chisari2021learning} indicated with $\dagger$ which uses a smaller form factor of 1:43 and would then reach a max speed over \SI{12}{\metre \per \second} when normalized at a 1:10 scale. }
\end{table}

\section{Methodology} \label{sec:methodology}
In this section, the hardware platform is introduced, while the {\namealgo} algorithm is presented in the following subsection. Finally, the training procedure and the simulator environment are described.
\subsection{Preliminaries} \label{ssec:preliminaries}
This work is developed and evaluated on a \gls{cots} F1TENTH car \cite{okelly2020f1tenthsim}, using an Intel i7-10710U as a computational platform.
The vehicle is set up for receiving steering angle and velocity commands as input, denoted with $\mathbf{u}=[\delta,\, v]^\top$, which are then translated into actuator inputs by the \gls{vesc} 6 MkIV.
All the experiments and evaluations on the real car use the localization and state estimation modules as defined in \cite{forzaeth}. 
For more details on the hardware and software components, and characterization of the localization pipeline used in this work, refer to \cite{forzaeth}.
Furthermore, two coordinate frames are used in this work: one is the traditional Cartesian coordinate frame and the second one is the curvilinear Frenet frame, in which coordinates are defined based on the reference line, using the $\left[s,\,d\right]$ vector, where $s$ generally denotes the advancement with respect to an arbitrary starting point on the path, and $d$ denotes the orthogonal distance from the same path. For further information on the reference frames and coordinate systems, also see \cite{forzaeth}.

\subsection{{\namealgo} Architecture} \label{ssec:res_ctrl_arch}
This section presents {\namealgo}, which comprises by the summation of two control actions, a baseline component $\mathbf{u_{PP}}$ and an \gls{rl} component $\mathbf{u_{RL}}$, resulting in the final input action 
\begin{equation} \label{eq:control_structure}
\mathbf{u} = \mathbf{u_{PP}} + \mathbf{u_{RL}}.
\end{equation}

The first part, the baseline controller, is a \gls{pp} controller, based on the work in \cite{purepursuit}. To summarize, this controller 
handles the steering dynamics of the vehicle by fixing a lookahead distance $d_{la}$, which is then used to select the lookahead point $\mathbf{p}$ on a predetermined path. 
\begin{comment}
% too long, commented
Assuming such a point will be reached along a constant curvature circle with radius $R$, the following equation can then be derived:
\begin{equation}
    \label{eq:pp_la}
    R = \frac{d_{la}^2}{2p_x},
\end{equation}
where $\mathbf{p} = (p_x,\,p_y)$ defines the lookahead point in the local frame relative to the car's rear-axle center.
Furthermore, assuming a kinematic bicycle model with no slip, the steering angle $\delta$ can be defined as a function of the radius of curvature $R$ along which the car is moving when such $\delta$ is kept constant:
\begin{equation}
    \label{eq:kin_model}
    \delta = \arctan \left(\frac{l_{wb}}{R}\right),
\end{equation}
where $l_{wb}$ refers to the vehicle's wheelbase length.
Hence, joining \Cref{eq:pp_la} and \Cref{eq:kin_model}, the \gls{pp} guidance law can be derived:
\begin{equation}
    \label{eq:pp_final}
    \delta = \arctan \left( \frac{2l_{wb}p_x}{d_{la}^2} \right)
\end{equation}
\end{comment}
Such a point is then assumed to be reachable along a constant curvature circle by assuming no tire slip, and the \gls{pp} guidance law hence derived reads as follows:
\begin{equation}
    \label{eq:pp_final}
    \delta = \arctan \left( \frac{2l_{wb}p_x}{d_{la}^2} \right)
\end{equation}
where $p_x$ is the lateral distance from the lookahead point $\mathbf{p}$ anf $l_{wb}$ is the car's wheelbase.
Notice that the no-slip assumption is quite simplistic, as a minimum amount of tire slip always happens when in corners \cite{PACEJKA20121}, and depending on the tires this could be significant even at slow speed. 

Longitudinally, the car's velocity is feed-forward controlled with $v_{PP}$, looking up the velocity profile generated by the trajectory generation optimizer at the closest point, as defined in \Cref{ssec:preliminaries}. 
A tuning parameter $\alpha_v$ is then used to linearly scale the velocity input, to account for the fact that at the highest velocities, the \gls{pp} model encounters great mismatch and is therefore unable to steer the car within the boundaries of a track. 
The eventual velocity input is therefore $v_{PP}=\alpha_v v_{ref}$, where $v_{ref}$ is the velocity obtained from the reference trajectory. 

The final \gls{pp} input is then named $\mathbf{u_{PP}} = \left[\delta_{PP} ,\, v_{PP}\right]^\top$, where $\delta_{PP}$ is obtained from \Cref{eq:pp_final}.
Two main tuning parameters are then retained, the first one, $d_{la}$, allows tuning the steering behavior: with this parameter lower values can improve tracking but introduce oscillations, and conversely, higher values improve control smoothness but could induce worse tracking and introduce both understeer and oversteer depending on the corner type.
The second one is the velocity gain $\alpha_v$ which can be selected to increase the reference velocity, at the cost of encountering more and more model mismatches due to effects such as tire slip.

The second part of the hybrid input is $\mathbf{u_{RL}}$, the outcome of a \gls{nn} that is trained as described in \Cref{ssec:rl_training}. This component is also scaled by a tuning factor, namely $\mathbf{u_{RL}} = \alpha_{RL}\,\mathbf{u_{NN}}$ where $\mathbf{u_{NN}}$ is the two-dimensional output of the policy network trained with \gls{rl}.

\subsection{Reinforcement Learning Training Procedure}
\label{ssec:rl_training}
\subsubsection{Simulator}
\label{ssec:sim}
The training procedure for the \gls{rl} agent is carried out in the F1TENTH simulator \cite{okelly2020f1tenthsim}. The model used in the simulator is the single-track vehicle model. The states of the model are the x-coordinate $X$, y-coordinate $Y$, heading angle $\varphi$, longitudinal velocity $v_x$, lateral velocity $v_y$, and yaw rate $r$. The inputs to the model are obtained from the combined input $\mathbf{u_{RL}}$, which directly provides reference steering angle $\delta$ and speed, which is then used to obtain an acceleration command $a$ with a P controller internal to the simulator.
\begin{align}
\dot X &=v_x \cos\varphi-v_y \sin \varphi \\
\dot Y &=v_x \sin\varphi+v_y \cos \varphi \\
\dot \varphi &=r \\
\dot v_x &=a+\frac{1}{m}(-F_{yf}\sin \delta+m v_y r) \\
\dot v_y &=\frac{1}{m}(F_{yr}+F_{yf}\cos \delta-m v_x r) \\
\dot r &=\frac{1}{I_z}(F_{yf}l_f\cos \delta-F_{yr}l_r) 
\end{align}
where \( m \) denotes the mass of the vehicle model and \( I_z \) represents the moment of inertia around the yaw axis. The distances from the center of gravity (CoG) to the front and rear axles are \( l_f \) and \( l_r \), respectively. Furthermore, \( F_{yf} \) and \( F_{yr} \) correspond to the lateral forces acting on the front and rear tires.

The tire forces are modeled using the Pacejka formula~\cite{PACEJKA20121}, as shown below:
\begin{align}
F_{y,i}=&\mu F_{z,i}D_i \sin  [C_i  \notag\\ &\arctan\left(B_i \alpha_i-E_i \left(B_i \alpha_i-\arctan\left(B_i \alpha_i\right) \right) \right) ]
\end{align}
where \( \mu \) represents the coefficient of friction, and \( F_{zi} \) denotes the load on the tire. The parameters \( B_i \), \( D_i \), \( C_i \), and \( E_i \) are tire-specific constants that need to be identified. The variable \( \alpha_i \) refers to the tire's sideslip angle.
\begin{align}
\alpha_f&=\arctan\left(\frac{v_y+ r l_f}{v_x}\right)-\delta \\
\alpha_r&=\arctan\left(\frac{v_y- r l_r}{v_x}\right)
\end{align}
The parameters used for this work are described in \Cref{tab:vehicle_parm}.

The tire friction $\mu$ was randomized with additive Gaussian noise with the distribution $\mathcal{N}(0,\,0.15)$, as previous work suggest \cite{Ghignone2023tcdriver, kaufmann2022benchmark}.
\begin{table}[htb]
        \centering
        \begin{tabular}{cccc}
        \toprule
        \textbf{ Parameters}& \textbf{Value}& \textbf{ Parameters}&\textbf{Value}\\
        \midrule
        $m$&           $3.56$ kg&                   $B_{f/r}$&           $7.67,20.00 $\\
        $I_{z}$&       $0.0627$ kg$\cdot$m$^{2}$&     $C_{f/r}$&           $0.48, 1.50$\\
        $l_f$&         $0.174$ m&                    $D_{f/r}$&           $2.00, 0.65$\\
        $l_r$&         $0.151$ m&                    $E_{f/r}$&           $1.10, 0.00$\\
        $\mu $&        $0.5$&                             &                    \\
        \bottomrule
        \end{tabular}
        \caption{Model Parameters}\label{tab:vehicle_parm}
\end{table}
\subsubsection{Reward}
The positive rewards utilized in this work consisted of a summation of normalized advancement $r_{adv}$ and normalized speed $r_{speed}$, to tradeoff between pure advancement on the track and higher instantaneous speed. Specifically, the two rewards are defined as follows, where time indices are omitted for ease of notation:
\begin{equation}
    r_{adv} = \frac{\Delta s}{v_{max}\,t_{sim}}, \quad r_{speed} = \frac{v}{v_{max}}, 
\end{equation}
where $v_{max}$ indicates the maximum velocity attainable in the simulator, $t_{sim}$ indicates the simulator timestep, $\Delta s$ indicates the current advancement in the Frenet $s$ coordinate with respect to the previous simulation iteration, and $v$ indicates the current speed of the car.
The penalties (negative rewards) consist instead of the following components: lateral deviation $r_{dev}$, relative orientation $r_{heading}$, collision with the track boundaries $r_{coll}$. They are defined as follows:
\begin{align}
    r_{dev} &= -\alpha_{dev} \frac{\mathbb{1}_{\mathcal{T}_{dev}}(|d|)}{w_{track}(s)},\\
    &\mathcal{T}_{dev} = \mathbb{R} \setminus \left[-\tau_{dev},\,\tau_{dev}\right] \nonumber \\
    r_{heading} &= -\alpha_{heading} \frac{\mathbb{1}_{\mathcal{T}_{\psi}}(|\psi-\psi_{ref}(s)|)}{\psi_{max}},\\
    &\mathcal{T}_{\psi} = \mathbb{R} \setminus \left[-\tau_{\psi},\,\tau_{\psi}\right] \nonumber \\ 
    r_{coll} &= -\mathbb{1}_{collision}.
\end{align}
The first two negative rewards are similarly built: both start from a deviation measurement, either the lateral deviation from the reference line $d$, or the difference in orientation between the vehicle heading $\psi$ and the angle of the tangent to the reference line at the current $s$ coordinate $\psi_{ref}(s)$; then both deviation measures are only considered if they are bigger than a threshold with an indicator function that excludes values lower than $\tau_{dev}$ and $\tau_{\psi}$, respectively; furthermore, both are normalized, either with respect to the track width $w_{track}$ at the current Frenet $s$ coordinate or a fixed maximum heading $\psi_{max}$; finally a scaling parameter is applied, $\alpha_{dev}$ and $\alpha_{heading}$, respectively.
The collision penalty instead only activates when the car is in collision with the boundaries, and has no scaling factor. 
A further addition introduced in this work is a direct linear scaling of the negative rewards based on the positive rewards: this scaling helped the learning procedure by not excessively penalizing deviations at the start and therefore allowing the car more initial freedom of movement.
The total reward $r_{tot}$ is then constituted as follows:
\begin{align}
    r_{pos} &= r_{adv} + r_{speed}\\
    r_{tot} &= r_{pos} + r_{pos}\left(r_{dev} + r_{heading}\right) + r_{coll}.
\end{align}
The numerical values of all the fixed parameters are defined in \Cref{tab:params}.

\begin{table}[h]
    \centering
    \begin{tabular}{cccc}
        \toprule
        $\alpha_{dev}$ & $\tau_{dev}$ & $\alpha_{heading}$ & $\tau_{\psi}$ \\
        \midrule
        1 &  0.1 & 0.25 & 0\\
        \bottomrule
    \end{tabular}
    \caption{Numerical values of \gls{rl} training parameters.}
    \label{tab:params}
\end{table}
\subsubsection{Observation Space}
Similarly to \cite{Ghignone2023tcdriver}, the observation space is made up of the following components: a part describing the state of the car, namely lateral deviation from the reference line $d$, relative orientation to the reference line $\Delta\psi=\psi-\psi_{ref}$, longitudinal and lateral velocities in the car's local frame $v_x,\,v_y$, and yaw rate $r$; the second part is made up of a segment of the track in front of the car, more specifically the left bound, the right bound and the reference line. Discrete points are sampled at equally spaced $s$ coordinates. 
Denoting with $N$ the total number of waypoints, the trajectory component then can be represented as:
\begin{gather*}
    \mathbf{o_{traj}} = \\
    [\mathbf{p}_{ref}^0,\, \hdots,\,\mathbf{p}_{ref}^N,\,\mathbf{p}_{left}^0,\, \hdots,\,\mathbf{p}_{left}^N,\,\mathbf{p}_{right}^0,\, \hdots,\,\mathbf{p}_{right}^N]^\top
\end{gather*}
where $p_{ref}$ represents a two-dimensional point on the reference line, $p_{left}$ a point on the left boundary and $p_{right}$ a point on the right boundary.
The full observation is then the concatenation of the 5 state variables plus the $6N$-dimensional trajectory component. For ease of notation, the observation space is also denoted as follows:
\begin{equation}
    \mathbf{o_{RL}} = [d,\,\Delta \psi,\,v_x,\,v_y,\,r,\,\mathbf{o_{traj}}]^\top
    ,\quad \mathbf{o_{RL}}\in\mathbb{R}^{5+6N}.
\end{equation}
Throughout this work, $N=20$ was selected.

\subsubsection{\acrshort{rl} training}
To train the \gls{nn}, the \gls{sac} algorithm \cite{sac} was used, using the provided implementation from the Stable Baselines3 library \cite{stable-baselines3}: such algorithm was selected for this work as it previously demonstrated to be effective in the \gls{ar} domain \cite{chisari2021learning, Ghignone2023tcdriver}. 
The \gls{nn} architecture consisted of a \gls{mlp} with two hidden layers of $256$ neurons each, the learning rate was set to \num{3e-4}, the buffer size was set to \num{1e6} and all other \gls{sac} parameters were left to the default values. The training was performed for \num{2e6} simulated steps, corresponding to shortly less than \SI{4}{\hour} on a workstation equipped with an Intel i7-13700K CPU and NVIDIA RTX 3090 GPU. 
Furthermore, as suggested in different previous sources, domain randomization was employed on the tire friction coefficient $\mu$, with more detail later explained in \Cref{ssec:sim}.
Finally, the initialization strategy of the velocity was modified, taking inspiration from Curriculum \gls{rl}, according to which presenting the agent with increasingly difficult environments can help with faster convergence and higher performance \cite{song2021overtaking, rudin2022learning}.
The initial velocity for the agent is randomly sampled at the beginning of every training episode around the average velocity of the previous episode: this effectively lets the agent start at slower speeds and increases the initial position only when the agent is consistently managing to achieve fast lap times. The specific distribution used was a Gaussian centered around the average velocity and with a \qty{0.5}{\metre \per \second} standard deviation.

\subsection{Comparison Controllers}
Within this work, the performance of multiple autonomous racing controllers is evaluated. These controllers serve as baselines to contextualize the performance of the proposed {\namealgo} controller:

\begin{enumerate}[I]
    \item \textbf{TC-Driver:} The TC-Driver is a trajectory-conditioned \gls{rl} agent, demonstrating superior generalization to various track conditions compared to a fully end-to-end agent \cite{Ghignone2023tcdriver}. However, its zero-shot transfer capability is still affected by model dynamics mismatch between simulation and reality (although this issue is less pronounced than in fully end-to-end learned agents) \cite{Ghignone2023tcdriver}. 
    
   \item \textbf{PP Controller:} The \gls{pp} controller \cite{purepursuit} operates under the non-slip assumption, adhering to a geometric control law as outlined in \Cref{ssec:res_ctrl_arch}. This assumption no longer holds true when the agent is pushed to the friction limits, resulting in lower racing performance compared to model-based control methods that incorporate tire dynamics \cite{map}.

    \item \textbf{MAP Controller:} The \gls{map} controller \cite{map} extends the \gls{pp} controller by utilizing a \gls{lut} to integrate the non-linear tire dynamics into the control action computed by the \gls{pp} controller. This \gls{sota} controller has demonstrated high effectiveness while maintaining computational and operational simplicity \cite{map, forzaeth}.
    
    \item \textbf{MPC:} The \gls{mpc} implementation follows a tracking approach inspired by \cite{hierarchicalmpc}, employing a single-track bicycle model with full non-linear tire dynamics as described in \cite{hierarchicalmpc}, with the objective of tracking the racing line. The \gls{mpc} computes an optimal control sequence within a receding horizon and achieves \gls{sota} racing performance while ensuring constraint satisfaction.

\end{enumerate}
All the algorithms from this work, both comparison ones and the presented {\namealgo}, use for reference a race line obtained employing the minimum curvature optimizer presented in \cite{Heilmeier2020MinimumCar}, which also provides a velocity profile on top of the positional coordinates. 
% Throughout this paper, \emph{racing line} and \emph{reference line} are interchangeably used to indicate the line provided by the said optimizer.

\begin{figure*}[!h]
\centering
\begin{subfigure}{.2\textwidth}
  \centering
  \includegraphics[width=\textwidth]{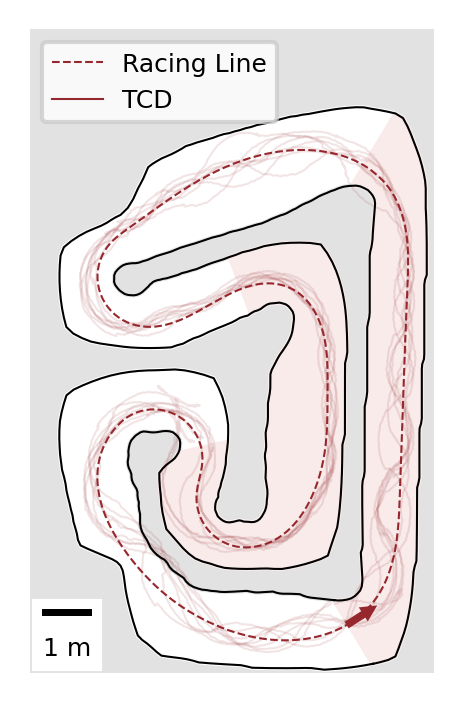}
  \caption{TC-Driver trajectories \textdagger}
  \label{fig:traj_tcd}
\end{subfigure}%
\begin{subfigure}{.2\textwidth}
  \centering
  \includegraphics[width=\textwidth]{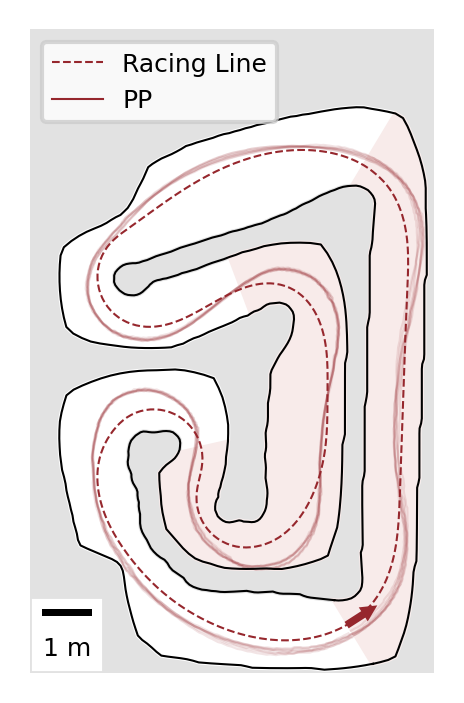}
  \caption{\gls{pp} trajectories}
  \label{fig:traj_pp}
\end{subfigure}%
\begin{subfigure}{.2\textwidth}
  \centering
  \includegraphics[width=\textwidth]{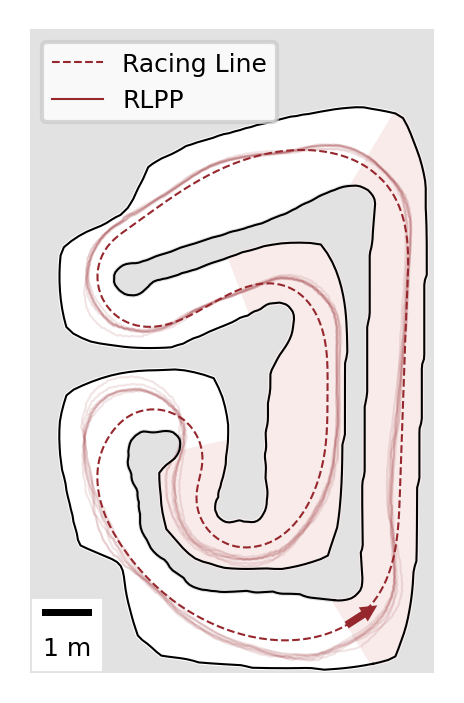}
  \caption{{\namealgo} trajectories}
  \label{fig:traj_rlpp}
\end{subfigure}%
\begin{subfigure}{.2\textwidth}
  \centering
  \includegraphics[width=\textwidth]{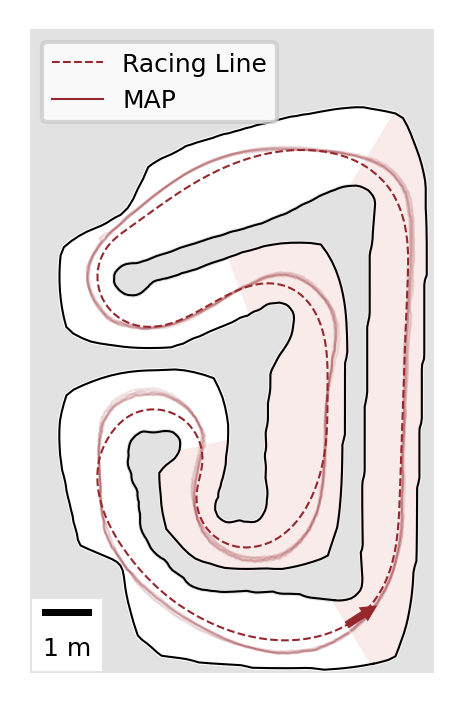}
  \caption{\acrshort{map} trajectories}
  \label{fig:traj_map}
\end{subfigure}%
\begin{subfigure}{.2\textwidth}
  \centering
  \includegraphics[width=\textwidth]{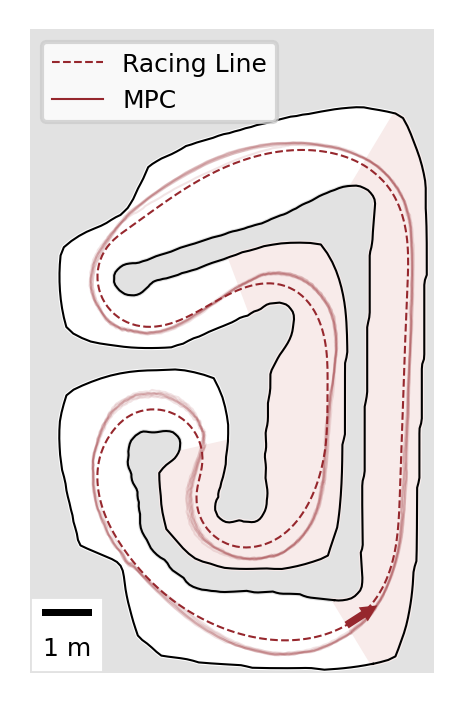}
  \caption{\gls{mpc} trajectories}
  \label{fig:traj_mpc}
\end{subfigure}%
\caption{Trajectories of the main compared algorithms. Ten consecutive laps are shown for each controller, except for TC-Driver (\textdagger) which only attained three consecutive laps without boundary violations. The arrow indicates the starting position for the Frenet $s$ coordinate. The dashed line represents the reference line that is used for each algorithm. Shaded areas within the track indicate 10-meter segments along the reference line, to facilitate references with \Cref{fig:vel_comp_pp}. The comparison between the comparison \gls{rl} method TC-Driver (a) and the proposed method (c) shows that the proposed method follows the reference line more smoothly.}
\label{fig:trajectories}
\end{figure*}

\section{Experimental Results} \label{sec:exp_results}
This section describes the experiments on the physical platform, and how the tuning procedure was carried out. Furthermore, an analysis on the simulation results and on the gap from reality is carried out.
\subsection{Tuning Procedure}
All the algorithms considered for the comparison were tuned on the real platform with the target of minimizing the lap time as much as possible while allowing for ten consecutive laps.  
As described in \Cref{ssec:res_ctrl_arch}, two main parameters are considered for tuning \gls{pp}: the lookahead distance $d_{la}$ and the velocity gain $\alpha_v$. 
The velocity gain is set as high as possible while avoiding crashes with the boundaries; the lookahead distance is increased if the behavior is oscillatory and decreased if low lateral tracking is achieved across the whole lap. The final parameters used for both {\namealgo} and \gls{pp} were $d_{la} = 1.2,\,\alpha_v=0.8$.
After such tuning is performed, the velocity gain parameter is transferred to the simulator, while the lookahead distance is not: as the simulated model dynamics are quite imprecise, transferring the same parameter does not lead to a well-behaved agent, and repeated crashes are sustained. 
Empirical assessments highlighted the need for reducing the initial speed of the \gls{pp} algorithm in simulation, in order to speed up initial learning, leading to a simulator value of $\alpha_v=0.75$.
Then the \gls{rl} agent is trained in simulation, using the hybrid control structure defined in \Cref{eq:control_structure}.
After training, the \gls{rl} agent is deployed on the real platform and again model mismatch is encountered: for this reason, the final value of $\alpha_{RL}$ is set to $0.55$.
Such amount is the larger amount that allows for 10 uninterrupted laps while improving the lap time when compared to the baseline \gls{pp} algorithm.
Notice that in this way {\namealgo} allows for simple tuning of one single parameter to alleviate the \sim2real gap, as opposed to complete retraining with different physics parameters.
A similar value of $0.55$ is then further used to reduce the velocity input of TC-Driver, as the agent would otherwise not be able to complete a single lap.

\subsection{Real Platform Experiments}
\label{ssec:real_experiments}
\sisetup{
round-mode = places,
round-precision = 2
}%
\begin{table}[hb]
    \vspace{-0.25cm}
    \centering
    \resizebox{\columnwidth}{!}{%
    \begin{tabular}{lcccccccc}
        \toprule
         & $\Bar{t}$ & $\sigma_t$ & $t_{min}$ & $t_{max}$ & $\Bar{d}$ & $\sigma_{d}$ & $\Bar{t}_{CPU}$ & $\sigma_{CPU}$ \\
         & $[s]$ & $[s]$ & $[s]$ & $[s]$ & $[m]$ & $[m]$ & $[ms]$ & $[ms]$ \\
         % & $\Bar{t}\;[s]$ & $\sigma_t \;[s]$ & $t_{min}\;[s]$ & $t_{max}\;[s]$ & $t_{CPU}\;[ms]$ & $\sigma_{CPU}\;[ms]$ \\
         \toprule
        TC-Driver$\dagger$ & \num{ 22.427273810000003 } & \num{ 1.0706559637937536 } & \num{ 21.131116867 } & \num{ 23.498123407 } & \textbf{\num[text-series-to-math]{ 0.15231246484175515 }} & \textbf{\num[text-series-to-math]{ 0.19491508195277502 }} & \num{ 7.442779775877907 } & \num{ 2.3790875689523343 } \\
        \gls{pp} & \num{ 14.3464390757 } & \textbf{\num[text-series-to-math]{ 0.042460159598790015 }} & \num{ 14.283079624 } & \num{ 14.429001809 } & \num[text-series-to-math]{ 0.2135580738236048 } & \num[text-series-to-math]{ 0.23529451083692257 }  & \textbf{\num[text-series-to-math]{ 1.6794986480197829 }} & \textbf{\num[text-series-to-math]{ 0.14911194393609453 }} \\
        \textbf{\namealgo} & \textbf{\num[text-series-to-math]{ 13.602260303399998 }} & \num[text-series-to-math]{ 0.1903690766846939 } & \textbf{\num[text-series-to-math]{ 13.366472244 }} & \textbf{\num[text-series-to-math]{ 13.942465782 }} & \num{ 0.21453208219054332 } & \num{ 0.256874562348986 } & \num{ 8.379565556176814 } & \num{ 2.7823980747449677 }
 \\
         \midrule
        \acrshort{map} & \textbf{\num[text-series-to-math]{ 12.6049916267 }} & \textbf{\num[text-series-to-math]{ 0.04687788034574503 }} & \textbf{\num[text-series-to-math]{ 12.558739423 }} & \textbf{\num[text-series-to-math]{ 12.686054707 }} & \textbf{\num[text-series-to-math]{ 0.14938052702389035 }} & \num[text-series-to-math]{ 0.1736785772096254 } & \textbf{\num[text-series-to-math]{ 1.8243235536901243 }} & \textbf{\num[text-series-to-math]{ 0.2634789071437146 }} \\
        \gls{mpc}& \num{ 12.747317171 } & \num{ 0.11286020525069726 } & \num{ 12.580536604 } & \num{ 12.992530584 } & \num{ 0.15662803699190736 } & \textbf{\num[text-series-to-math]{ 0.1704082139282545 }} & \num{ 11.231001493003584 } & \num{ 2.8745781973911875 } \\
         \bottomrule
    \end{tabular}
    }
    \caption{Lap times statistics of ten consecutive laps around the reference track shown in \Cref{fig:trajectories}. Mean and standard deviation are indicated with $\Bar{t},\,\sigma_t$, respectively, and minimum and maximum lap time are indicated with $t_{min},\,t_{max}$. Mean and standard deviation of the absolute lateral deviation are reported with $\Bar{d},\,\sigma_d$. Mean and standard deviation of the computation times are denoted by $\Bar{t}_{CPU},\,\sigma_{CPU}$. $\dagger$: TC-Driver was not able to complete 10 consecutive laps, therefore the statistics presented here are computed across 4 laps. The middle line separates the first three tire-free methods from the latter two \gls{sota} methods incorporating tire models.}
    \label{tab:laptimes}
\end{table}
% \begin{table}[hb]
%     \vspace{-0.25cm}
%     \centering
%     \resizebox{\columnwidth}{!}{%
%     \begin{tabular}{lcccc}
%         \toprule
%          & $\Bar{t}$ & $\sigma_t$ & $t_{min}$ & $t_{max}$ \\
%          & $[s]$ & $[s]$ & $[s]$ & $[s]$ \\
%          % & $\Bar{t}\;[s]$ & $\sigma_t \;[s]$ & $t_{min}\;[s]$ & $t_{max}\;[s]$ & $t_{CPU}\;[ms]$ & $\sigma_{CPU}\;[ms]$ \\
%          \toprule
%         TC-Driver & \num{ 22.427273810000003 } & \num{ 1.0706559637937536 } & \num{ 21.131116867 } & \num{ 23.498123407 } \\
%         \gls{pp} & \num{ 14.3464390757 } & \textbf{\num[text-series-to-math]{ 0.042460159598790015 }} & \num{ 14.283079624 } & \num{ 14.429001809 } \\
%         \textbf{\namealgo} & \textbf{\num[text-series-to-math]{ 13.602260303399998 }} & \num[text-series-to-math]{ 0.1903690766846939 } & \textbf{\num[text-series-to-math]{ 13.366472244 }} & \textbf{\num[text-series-to-math]{ 13.942465782 }} 
%  \\
%          \midrule
%         \acrshort{map} & \textbf{\num[text-series-to-math]{ 12.6049916267 }} & \textbf{\num[text-series-to-math]{ 0.04687788034574503 }} & \textbf{\num[text-series-to-math]{ 12.558739423 }} & \textbf{\num[text-series-to-math]{ 12.686054707 }}\\
%          \bottomrule
%     \end{tabular}
%     }
% \end{table}
The presented {\namealgo} was tested on the real car on a fixed racing track, shown in \Cref{fig:trajectories}.
All the algorithms were tested on the same track, after a tuning procedure that aimed at minimizing lap time while still allowing 10 consecutive laps to be run without collisions with the track boundaries. 
The lap time results can be seen in \Cref{tab:laptimes}: comparing the proposed {\namealgo} to the baseline \gls{pp} controller our method improves lap time by \qty{5.23}{\percent}, going from \qty{14.35}{\second} to \qty{13.60}{\second}. The baseline \gls{pp} controller turns out more consistent in the resulting lap times, but this results even in a greater improvement of \qty{6.37}{\percent} when considering $t_{min}$, arguably an even more important metric in \gls{ar} compared to average lap time.
When considering the comparison controllers with tire-model knowledge, both \gls{map} and \gls{mpc} are faster than the method presented here, respectively by \qty{7.35}{\percent} and \qty{7.27}{\percent} when considering average lap time; however, {\namealgo} closes the gap from the model-free \gls{pp} considerably, for instance, considering the minimum lap time, {\namealgo} closes the gap by \qty{52.9}{\percent} against \gls{map} and by \qty{53.5}{\percent} against \gls{mpc}.

\begin{figure}[h]
    \centering
    \includegraphics[trim={0cm 1.11cm 0cm 0cm}, clip]{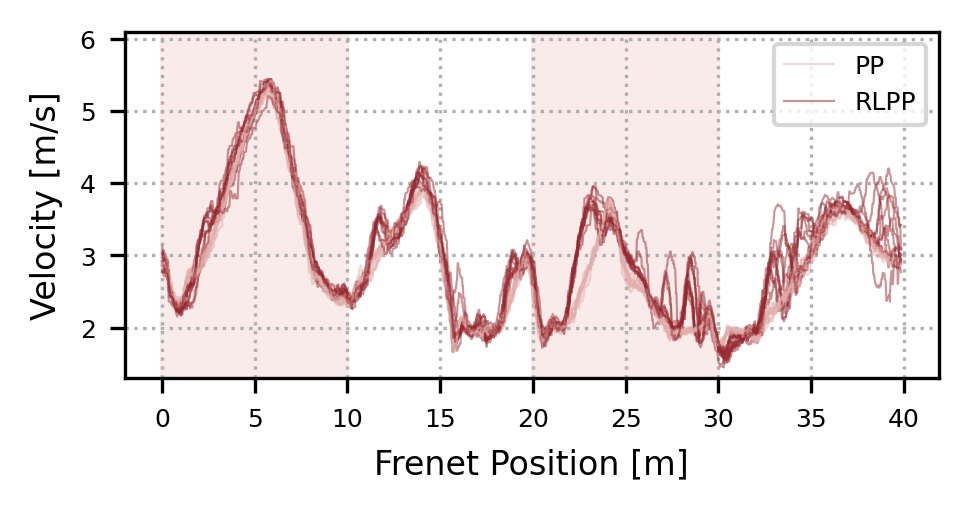}
    \includegraphics[trim={0cm 0.1cm 0cm 0.2cm}, clip]{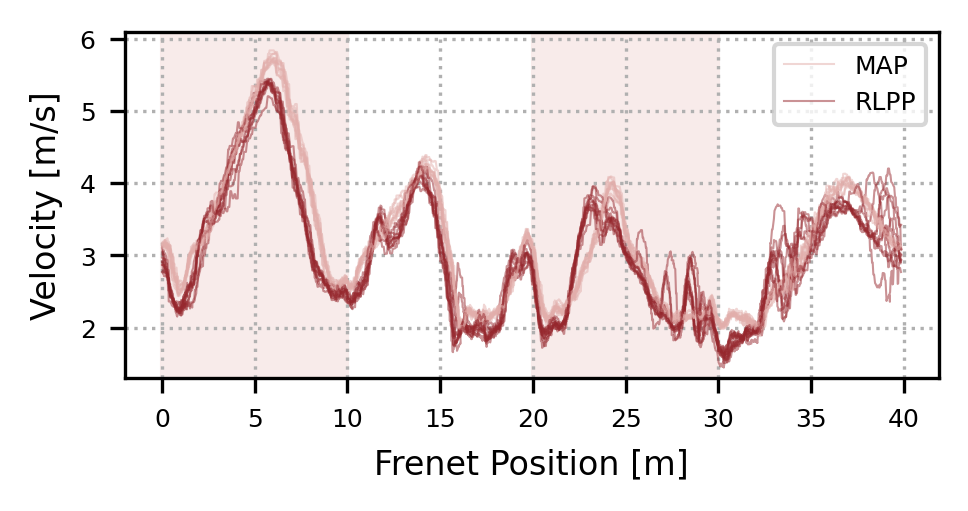}
    \caption{Velocity profile comparison across ten consecutive laps, plotted against the Frenet $s$ coordinate. The algorithm presented in this paper is in the darker hue, while \gls{pp} (top) and \gls{map} (bottom) are shown with the lighter hue. Shaded areas are present to facilitate references with \Cref{fig:trajectories}.}
    \label{fig:vel_comp_pp}
\end{figure}

To better understand where the differences between the controllers arise, the different trajectories are presented in \Cref{fig:trajectories} and a comparison of the velocity profiles of {\namealgo} with \gls{pp} and \gls{map} are present in \Cref{fig:vel_comp_pp}.
From the velocity profile, it can be seen the first main modification of {\namealgo} to the baseline behavior comes at the beginning of the long acceleration phases in short straights: it is visible that after the 10-, 20-, and 30-meter mark the \gls{rl} strategy tends to increase velocity. A correspondence can then be found in \Cref{fig:trajectories}, as these three parts are either short straight segments of the track (after the 10 and 20-meter marks) or with mild curvature (after the 30-meter mark). It is then interesting to notice the significantly more marginal difference in the main straight, starting after the zero-meter mark, where {\namealgo} modifies significantly less the acceleration behavior. 
The second component of {\namealgo} can then be witnessed in the steering behavior, which can be analyzed comparing \Cref{fig:traj_pp} and \Cref{fig:traj_rlpp}, where \gls{pp} and {\namealgo} are shown, respectively.
It is evident, specifically after the 0- and the 20-meter marks, that our introduced algorithm improves the straight-line behavior, yielding a trajectory generally more aligned with the reference line. However, this comes at the expense of lateral error, as it is specifically visible in the chicane straddling the 30-meter line, where {\namealgo} struggles more to complete the second corner. 
Comparing then the full behavior of our proposed solution with \gls{map}, it becomes clearer where the lack of performance arises from. 
Firstly, the integrated model knowledge enables \gls{map} to track the trajectory much better, as can be seen in \Cref{fig:traj_map} and \Cref{tab:laptimes}.
Furthermore, this better steering handling is coupled with a higher velocity profile, which can be seen in \Cref{fig:vel_comp_pp}: specifically around 5 and 35 meters \gls{map} manages to handle higher top speeds, and ultimately achieve a lower lap time.
Finally, when it comes to computation, it is clear that {\namealgo} is the highest method within the model-free ones, but the required time of \qty{8.38}{\milli \second} is still low enough to allow for the required \qty{40}{\hertz} frequency. 
% \begin{figure}
%     \centering
%     \includegraphics{fig/vel_comp/vel_comparison_map_rlpp.png}
%     \caption{Velocity profile comparison across ten consecutive laps, plotted against the Frenet $s$ coordinate. The algorithm presented in this paper is in the darker hue, while \acrshort{map} is shown with the lighter hue. Shaded areas are present to facilitate references with \Cref{fig:trajectories}.}
%     \label{fig:vel_comp_map}
% \end{figure}

\subsection{\sim2real Gap}
To compare the performance difference of the \gls{rl} algorithms from simulation to reality, the final settings that were used in the physical setting were replicated in simulation and reported in \Cref{tab:sim2real}.
As it can be noticed, not only {\namealgo} maintains the inferior lap times when compared to the baseline TC-Driver and to the tire-free \gls{pp}, but also manages to achieve the overall smallest \sim2real gap where this is computed as the percentual difference of the simulation time with respect to the real-time.

% \sisetup{
% round-mode = places,
% round-precision = 2
% }%
% \begin{table}[b]
%     \centering
%     \begin{tabular}{lcccccc}
%          & $\Bar{t}$ & $\sigma_t$ & $t_{min}$ & $t_{max}$ & $\Bar{d}$ & $\sigma_{d}$\\
%          & $[s]$ & $[s]$ & $[s]$ & $[s]$ & $[m]$ & $[m]$\\
%          % & $\Bar{t}\;[s]$ & $\sigma_t \;[s]$ & $t_{min}\;[s]$ & $t_{max}\;[s]$ & $t_{CPU}\;[ms]$ & $\sigma_{CPU}\;[ms]$ \\
%          \toprule
%          {\namealgo} (simulation) & \num[text-series-to-math]{ 13.614999999999998 } & \textbf{\num[text-series-to-math]{ 0.15173990905493545 }} & \num[text-series-to-math]{ 13.45 } & \textbf{\num[text-series-to-math]{ 13.85 }} & \textbf{\num[text-series-to-math]{ 0.05357874406315403 }} & \textbf{\num[text-series-to-math]{ 0.049836678465732455 }} \\
%         {\namealgo} (real car) & \textbf{\num[text-series-to-math]{ 13.602260303399998 }} & \num[text-series-to-math]{ 0.1903690766846939 } & \textbf{\num[text-series-to-math]{ 13.366472244 }} & \num[text-series-to-math]{ 13.942465782 }  & \num{ 0.21453208219054332 } & \num{ 0.256874562348986 }\\
%          \bottomrule
%     \end{tabular}
%     \caption{Lap times statistics of ten consecutive laps around the reference track for the same {\namealgo} network, both in simulation and on the physical platform. Mean and standard deviation are indicated with $\Bar{t},\,\sigma_t$, respectively, and minimum and maximum lap time are indicated with $t_{min},\,t_{max}$. The mean and standard deviation values of the absolute lateral deviation are also reported with $\Bar{d},\,\sigma_d$.}
%     \label{tab:laptimes_sim}
% \end{table}

\sisetup{
round-mode = places,
round-precision = 3
}%
\begin{table}[h]
    \centering
    \begin{tabular}{lccc}
        \toprule
         & TC-Driver & \gls{pp} & {\namealgo} \\
         \toprule
         $\Bar{t}$ (sim) $\downarrow$ &  \num[text-series-to-math]{ 18.5 } & \num[text-series-to-math]{ 14.95 } & \textbf{\num[text-series-to-math]{ 13.890 }} \\
         $\Bar{t}$ (real) $\downarrow$ &  \num{ 22.427273810000003 } & \num{ 14.3464390757 } & \textbf{\num[text-series-to-math]{ 13.602260303399998 }} \\
         \sim2real gap $\downarrow$ & \qty{17.5111511}{\percent} & \qty{4.207043442036509}{\percent} & \textbf{\qty[text-series-to-math]{2.115381489413784}{\percent}} \\
         \bottomrule
    \end{tabular}
    \caption{\sim2real comparison from data averaged across ten laps. The three controllers here considered were modified in simulation to use the same parameters as in real, as detailed in \Cref{ssec:real_experiments}}
    \label{tab:sim2real}
\end{table}
\sisetup{
round-mode = places,
round-precision = 2
}%

\section{Conclusion} \label{sec:conclusion}
This paper presents and evaluates a zero-shot \gls{rl} method deployed on an autonomous racing platform that surpasses \gls{sota} model-free controllers by leveraging a residual architecture. 
The proposed algorithm manages to outperform the baseline \gls{pp} controller by \qty{6.37}{\percent} when considering minimum lap times, and closes the gap to the overall best-performing model-based methods by \qty{52.90}{\percent}.
The performance increase comes for only a fraction of increased computational requirements, reaching a total computation time of \qty{8.38}{\milli \second}, way below the threshold for running the controller at the requested \qty{40}{\hertz}.

One direction for future work includes how the method improves complex controllers that use tire dynamics, such as combining residuals with \gls{map} or \gls{mpc}. Another direction is to close the \sim2real gap by enhancing the simulator with data-driven methods or training \gls{rl} on the real car. 

% \section*{Acknowledgement}
% The authors would like to thank all team members of the
% \textit{ForzaETH} racing team.

%%%%%%%%%%%%%%%%%%%%%%%%%%%%%%%%%%%%%%%%%%%%%%%%%%%%%%%%%%%%%%%%%%%%%%%%%%%%%%%%
\bibliographystyle{IEEEtran}
\bibliography{main}

\end{document}

%% file: acronyms.tex
% \newacronym{s2r}{Sim-to-Real}{Simulation-to-Reality}
\newacronym{rl}{RL}{Reinforcement Learning}
\newacronym{mpc}{MPC}{Model Predictive Control}
\newacronym{pp}{PP}{Pure Pursuit}
\newacronym{map}{MAP}{Model- and Acceleration-based Pursuit}
\newacronym{sota}{SotA}{State-of-the-Art}
\newacronym{ar}{AR}{Autonomous Racing}
\newacronym{ml}{ML}{Machine Learning}
\newacronym{nn}{NN}{Neural Network}
\newacronym{vesc}{VESC}{Vedder Electronic Speed Controller}
\newacronym{cots}{CotS}{Commercial off-the-Shelf}
\newacronym{sac}{SAC}{Soft Actor-Critic}
\newacronym{mlp}{MLP}{Multilayer Perceptron}
\newacronym{lut}{LuT}{Lookup Table}

%% file: keywords.tex
\def\namealgo{RLPP}
\def\sim2real{Sim-to-Real}